%% file: 00_kg_bias_workshop_submission.tex
\documentclass[sigconf
]{acmart}

\usepackage{xcolor}
\usepackage{todonotes}
\usepackage{multirow}

\usepackage{graphicx}
\usepackage[utf8]{inputenc} %
\usepackage[T1]{fontenc}    %
\usepackage{hyperref}       %
\usepackage{url}            %
\usepackage{booktabs}       %
\usepackage{amsfonts}       %
\usepackage{nicefrac}       %
\usepackage{microtype}      %
\usepackage{amsmath}

\usepackage{subcaption}
\usepackage{makecell}

\usepackage{breqn}
\AtBeginDocument{%
  \providecommand\BibTeX{{%
    \normalfont B\kern-0.5em{\scshape i\kern-0.25em b}\kern-0.8em\TeX}}}

\makeatletter
\renewcommand\@formatdoi[1]{\ignorespaces}
\makeatother

\setcopyright{none}
\renewcommand\footnotetextcopyrightpermission[1]{}

\copyrightyear{2020}
\acmYear{2020}

\acmConference[KDD 2020]{International Workshop on
Mining and Learning with Graphs at KDD Conference}{Aug 24, 2020}{San Diego, CA}
\acmBooktitle{MLG 2020: 16th International Workshop on
Mining and Learning with Graphs - A Workshop at the KDD Conference, August 24, 2020, San Diego, CA}

\begin{document}

\title{Adversarial Learning for \\
Debiasing Knowledge Graph Embeddings}

\author{Mario Arduini}
\authornote{The first three authors contributed equally to this research.}
\email{maarduin@ethz.ch}
\affiliation{%
  \institution{ETH Z\"urich}
  \country{Switzerland}
}

\author{Lorenzo Noci}
\authornotemark[1]
\email{lnoci@ethz.ch}
\affiliation{%
  \institution{ETH Z\"urich}
  \country{Switzerland}
}

\author{Federico Pirovano}
\authornotemark[1]
\email{fpirovan@ethz.ch}
\affiliation{%
  \institution{ETH Z\"urich}
  \country{Switzerland}
}

\author{Ce Zhang}
\email{ce.zhang@inf.ethz.ch}
\affiliation{%
  \institution{ETH Z\"urich}
  \country{Switzerland}
}

\author{Yash Raj Shrestha}
\email{yshrestha@ethz.ch}
\affiliation{%
  \institution{ETH Z\"urich}
  \country{Switzerland}
}

\author{Bibek Paudel}
\email{bibekp@stanford.edu}
\affiliation{%
  \institution{Stanford University}
  \country{U.S.A.}
}

\begin{abstract}
Knowledge Graphs (KG) are gaining increasing attention in both academia and industry. Despite their diverse benefits, recent research and anecdotal evidence has identified societal and cultural biases embedded in the representations learned from KGs. 
Such biases may have detrimental consequences for minority groups in the population as the applications of KGs begin to intersect and interact with the social sphere.
This work aims at identifying and mitigating such biases in knowledge graph embeddings.
As a first step, we explore popularity bias, which refers to the relationship between node popularity and link prediction accuracy.
In case of node2vec graph embeddings, we find that the prediction accuracy of the embedding is negatively correlated with the degree of the node.
In contrast, we observe an opposite trend with Knowledge Graph Embeddings (KGE).
As a second step, we explore gender bias in KGE, and a careful examination of popular KGE algorithms suggests that sensitive attributes such as gender of a person can be predicted from the embedding.
This implies that such biases in popular KGs is captured by the structural properties of the embedding.
As a preliminary solution to debiasing KGs, we introduce a novel framework to filter out the sensitive attribute information from the KGE, which we call FAN (Filtering Adversarial Network). 
We also suggest the applicability of FAN for debiasing other network embeddings, which could be explored in future work.
\end{abstract}

\ccsdesc[300]{Information systems~Document representation}
\ccsdesc[300]{Information systems~Data mining}
\ccsdesc[500]{Computing methodologies~Machine learning}
\ccsdesc[500]{Computing methodologies~Knowledge representation and reasoning}

\keywords{network embedding, knowledge graph embedding, representation learning, bias, fairness}

\maketitle

\section{Introduction}
\input{01_introduction}

\label{sec:introduction}

\section{Biases in Network and Knowledge Graph Embeddings}
\input{02_bias_in_kg}
\label{sec:bias_in_kg}

\section{Debiasing Knowledge Graph Embedding}
\input{03_debiasing_method}
\label{sec:debiasing}

\section{Experiments and Results}
\input{04_experiments_results}
\label{sec:experiments}

\section{Limitations, Conclusion, and Future Work}
\input{05_conclusion}
\label{sec:conclusion}

\bibliographystyle{ACM-Reference-Format}
\bibliography{09_kg_bias_bibliography}

\end{document}

%% file: 01_introduction.tex
Multi-relational graphs, composed of entities (nodes) and edges representing semantic meaning, popularly known as {\it Knowledge Graphs (KG)}~\cite{wang2014knowledge}, are gaining increasing industrial applications. For instance, the Google search engine uses the \textit{Google Knowledge Graph} to facilitate linking semantic information from various websites in a unified view. Other applications of KGs include data governance, automatic fraud detection, and knowlege management. As a consequence, academic research on KGs both from the lens of machine learning and representation learning is gaining a lot of momentum. Research on machine learning on KGs identify a diverse set of inference techniques that can be applied on KGs, including logical rules mining \cite{ho2018rule, ho2018learning}, semantic parsing \cite{berant2013semantic, heck2013leveraging}, named entity disambiguation \cite{damljanovic2012named, zheng2012entity}, and information extraction \cite{bordes2014open, bordes2014question}. Research on KG representation learning aim to build useful representations for entities and relations with high reliability, explainability, and reusability. Representation learning on KGs is a very active line of research, with numerous novel Knowledge Graph Embedding (KGE) algorithms being proposed recently, including \textit{TransE} \cite{bordes2013translating}, \textit{TransD} \cite{ji2015knowledge}, \textit{TransH} \cite{wang2014knowledge}, \textit{RESCAL} \cite{nickel2011three}, \textit{DistMult} \cite{yang2014embedding}, \textit{HolE} \cite{nickel2016holographic}, \textit{CrossE} \cite{zhang2019interaction}, \textit{ComplEx} \cite{trouillon2016complex}, and \textit{Analogy} \cite{liu2017analogical}. Simultaneously, in the related field of network and graph representation learning, several advances have been made in the development of accurate graph embedding methods, including \textit{Deepwalk} \cite{perozzi2014deepwalk} and \textit{node2vec} \cite{grover2016node2vec}.

Together with these advances in embedding learning methods,  recent years has also witnessed various anecdotal evidences suggesting that these methods amplify biases already present in the data  \cite{DBLP:journals/algorithms/Shrestha019}. Empirical investigations have also identified biases embedded in knowledge graph representations. For instance, a recent article by Janowicz et al. \cite{Janowicz2018DebiasingKG} identified the existence of social biases in KGs. The existence of such biases is detrimental to the usability of the knowledge graph. This is especially true when KG applications such as search engines \cite{wang2014knowledge} and knowledge management systems are penetrating the social spheres. Besides a few exceptions \cite{DBLP:conf/www/Demartini19}, research works on the identification and mitigation of such social biases in KGs remain absent. The absence of coherent and useful debiasing frameworks for KGs is problematic, and could lead to detrimental societal consequences, in particular with respect to minorities. To tackle  this problematic gap in the literature, we aim to characterize, investigate, and develop methods for mitigating social biases that arise from network and knowledge graph embedding algorithms.

Our empirical exercise comprises of two elements. First, we examine simple networks, with unlabelled relations, and identify the existence of what we call a \textit{popularity bias}, i.e. a correlation between the popularity (degree) of the nodes and the link prediction accuracy of the embeddings. Research in recommendation systems have reported the presence of popularity biases in well-established ranking algorithms and ways to mitigate them. As network embeddings find use in downstream tasks like search and recommendations, it is important to study the presence and ways to mitigate such biases as well. Our findings suggest that structural information on low-degree nodes is captured more accurately than on high-degree nodes by popular network embeddings algorithms such as \textit{Deepwalk} \cite{perozzi2014deepwalk} and \textit{node2vec} \cite{grover2016node2vec}.
Second, we intend to characterise and mitigate the \textit{inference bias} that arises while training rules with classifiers operating in the KGE space \cite{Janowicz2018DebiasingKG}. As a result, we identify how some sensitive attributes, such as  gender, are captured by popular KGE algorithms, such as \textit{TransE} \cite{bordes2013translating}, \textit{TransD} \cite{ji2015knowledge}, and \textit{TransH} \cite{wang2014knowledge}. Additionally, we find  that the gender attribute is still captured in the embedding even when  gender relations are explicitly removed from the graph.

To summarize, our findings suggest that sensitive attributes in KGs are not only represented by explicit relations bearing the name of such attributes, but rather they are embedded in the structure of the whole graph structure. An important implication of this finding is the necessity of fine-grained debiasing algorithm operating \textit{on the embeddings}, instead of just removing the sensitive relations from the KG. As a useful solution, we develop a debiasing method based on adverserial learning that modifies the embeddings by filtering out sensitive information, while aiming to preserve all the other relevant information. With empirical examination, we show the applicability of our method in removing  gender bias in KGEs for both high-degree and low-degree entities.

We present our method and findings of this \textbf{work in progress} in the following order. In Section~\ref{sec:bias_in_kg} we introduce the embedding methods and the biases we aim to study. In Section~\ref{sec:debiasing}, we present our approach for debiasing KGEs, followed by experimental results in Section~\ref{sec:experiments}.  Conclusions and direction for future work is presented in Section~\ref{sec:conclusion}.

%% file: 02_bias_in_kg.tex
In this section, we briefly describe two different types of graph bias, and  present the embedding algorithms examined throughout the paper.
First, we provide a brief overview on networks and KGEs.

A network contains a set of nodes $N$ and edges $E \subseteq \{N \times N\}$ that encode relationships between the nodes.
KGs contain labeled relationships between entities in the form of triples $\langle h, r, t \rangle$, where $h$ is the head entity, $t$ is the tail entity, and $r$ is the relation between then. An example of a KG triple is $\langle Albert\_Einstein, born\_in, Ulm \rangle$.
Embedding learning algorithms aim at learning real-valued representations of nodes, entities, and relations in some low-dimensional space.
Specifically, network embeddings learn a vector $n_i \in \mathbb{R}^d$ for each node $i \in N$ in the network.
The dimension of the embedding is represented by $d$.
Similarly, KGEs learn embeddings for entities and relations.
Often, the embeddings are learned by training on an objective function that maximizes the probability of true edges and triples (those that exist in the training dataset), and they are evaluated by their performance on link prediction on the testing dataset.

\subsection{Popularity Bias in Network Embeddings}
We define popularity bias as the bias resulting from correlation between the degree of a node in a graph and the accuracy of link prediction of the embedding of the node.
In the recommendation systems literature, it has been reported that such biases lead to promotion of blockbuster items to the detriment of long-tail items, many of which could be interesting to the users~\cite{paudel2016updatable}. Since network embeddings are also increasingly used in search and recommendations, such biases could affect these downstream tasks and lead to lack of diversity and filter bubbles in users' online experiences.

To investigate whether network embeddings exhibit popularity bias, we examined the popular node2vec~\cite{grover2016node2vec} method on the benchmark AstroPh dataset. The AstroPh dataset represents the  network of collaborations between astrophysicists extracted from papers submitted to the e-print website arXiv. The nodes represent scientists, and an edge is present between two scientists if and only if they are listed as co-authors in at least one paper present in the repository. The network consists of $187,22$ nodes and $198,110$ edges.

Before describing node2vec, we briefly discuss the DeepWalk~\cite{perozzi2014deepwalk} algorithm, upon which node2vec is based. DeepWalk extracts latent representations from networks in the following way.
First, the algorithm iteratively builds a corpus of random walks for each node. Each random walk has a fixed length, and the next node in the walk is chosen at random among neighbouring nodes of the current node. Importantly, the same fixed number of random walks are calculated for each node, regardless of their degree.
Next, this corpus of random walks is fed into a \textit{SkipGram}~\cite{mikolov2013efficient} model to learn the latent representations.
The embeddings can then be used for downstream tasks, including link prediction and node classification.

The node2vec algorithm works in a very similar way to DeepWalk. The basic steps remain the same, i.e. the algorithm first builds a corpus of random walks for each node, and then this corpus is fed into a \textit{Skip-Gram} model in order to learn the embeddings. The only difference between the two methods is the way in which the random walks are explored. 
Instead of being sampled uniformly from the current node's neighbourhood (as in Deepwalk), the random walk traversal in node2vec is done using a parametric set of transition probabilities. This parametric form allows for a fine-grained and balanced tuning between the extreme sampling scenarios of Breadth-First Search (BFS) and Depth-First Search (DFS).

\begin{figure*}[ht!]
\includegraphics[width=\textwidth]{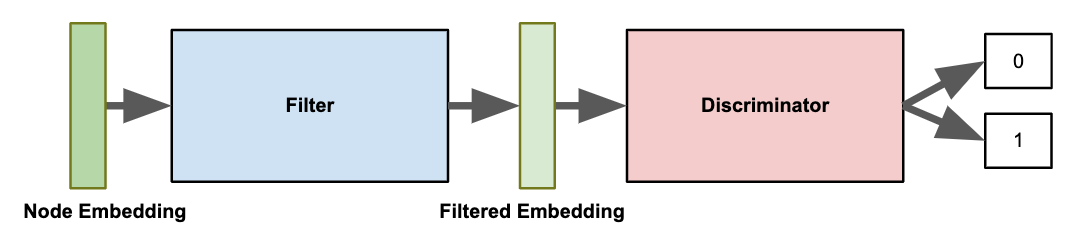}
\caption{FAN model: the filter takes as input a vector and outputs its filtered version (ideally without the sensitive attribute). The discriminator tries to predict the sensitive attribute (in the figure it is assumed to be binary) from the filtered embedding, and ideally will reach an accuracy of 50\% (random prediction).}
\label{fan_model}
\end{figure*}

\subsection{Gender Bias in KGE}
KGEs might exhibit several societal biases, such as ethnicity, gender, religion, etc.
We follow prior work in this area~\cite{caliskan2017semantics} to define the presence of such biases.
We intend to explore how different attributes interact in KGEs and to remove sensitive attributes (e.g., gender, ethnicity) from the embedding while preserving all other information. 
In this prelimilary work, we limit ourselves to the problem of gender bias and expect that our filtered embeddings do not correlate gender information with non-gender information.
To this end, we treat gender as a sensitive attribute and perform occupation prediction (which in our case is posed as an unbalanced multiclass classification problem) training a simple neural network operating in the embedding space. 
In this way, we can measure the interaction between the gender sensitive attribute and the occupation non-sensitive attribute, and use this information as evidence for the existence of bias.

Given its popular use and huge size, we adopt the \textit{DBPedia} \cite{auer2007dbpedia} dataset for our empirical investigation on KGEs. Based on scalability and simplicity of use, we focus our analysis on three popularly used KGE algorithms, namely  \textit{TransE}~\cite{bordes2013translating}, \textit{TransH}~\cite{wang2014knowledge}, and \textit{TransD}~\cite{ji2015knowledge}.
These algorithms have increasing complexity, leading to more powerful and data-savvy embeddings, at the cost of more computationally-expensive training.
For each of them we used the implementation provided by OpenKe \cite{han2018openke}.
These algorithms differ in the loss function used and in their number of parameters. We present a brief overview of the methods and their properties.

\subsubsection{TransE \cite{bordes2013translating}}
The basic principle behind TransE is the use of the translation operation to generate the embedding of a tail entity, given the embeddings for the head and the relation. 
It assigns one embedding to each node and one embedding to each relation.
TransE uses minibatch stochastic gradient descent to minimize a loss function on the embeddings for real triples present in the graph, while doing negative sampling to generate false triples and maximizing their loss.
The loss function $f_r(h,t) = \lVert \textbf{h} + \textbf{r} -\textbf{t} \rVert $ is the euclidean distance between the embedding of the tail and the embedding of the head plus the embedding of the relation.
The problem with this approach is evident in \textit{many-to-1} relations, for example gender, because in this case to minimize the loss for all gender triples, all persons (which are different nodes in the graph) that have the same gender are forced to have representations that are close in the embedding space.

\subsubsection{TransH \cite{wang2014knowledge}}
TransH overcomes the drawbacks of TransE by allowing an entity to have distinct representations when dealing with different relations, i.e. \textit{many-to-1} relations.
In order to make this possible, the authors introduced to the TransE framework an additional relation-specific vector $\textbf{w}_r$ to project the entities on an hyperplane with this vector as normal vector.
For the loss function, we calculate the projected head and tail as $\textbf{h}_{\perp} = \textbf{h} - \textbf{w}_r^{\top}\textbf{h}\textbf{w}$ and $\textbf{t}_{\perp} = \textbf{t} - \textbf{w}_r^{\top}\textbf{t}\textbf{w}_r$.
Then we calculate the loss as before as $f_r(h,t) = \lVert \textbf{h}_{\perp} + \textbf{r} -\textbf{t}_{\perp} \rVert $ and apply SGD using negative sampling.

\subsubsection{TransD \cite{ji2015knowledge}}
TransD works following the same principle as TransH. However, instead of using a projecting vector, it utilizes a projection matrix which can be decomposed as the identity matrix added to the outer product of two vectors, one that is relation-specific and another that is entity-specific.
The projection matrix is calculated as follows: $\textbf{M}_r^t = \textbf{w}_r\textbf{w}_t^{\top} + \textbf{I}$.
We then calculate $\textbf{h}_{\perp} = \textbf{M}_r^h\textbf{h}$ and $\textbf{t}_{\perp} = \textbf{M}_r^t\textbf{t}$ and the loss in the same way as we did for TransH.

As it is evident that both TransD and TransH are able to capture \textit{many-to-1} and \textit{many-to-many} relations way more effectively than TransE, we use TransH and TransD for our experiments.

%% file: 03_debiasing_method.tex
As a first solution for debiasing knowledge graph embeddings, we developed an adversarial model which we call FAN (Filtering Adversarial Network). The model is an adversarial network composed of two players, a filter module $F_{\theta_f}:\mathbb{R}^d \rightarrow \mathbb{R}^d$ that aims at filtering out the information about the sensitive attribute from the input, and a discriminator module $D_{\theta_d}:\mathbb{R}^d \rightarrow \left[0,1\right]$ that aims at predicting the sensitive attribute from the output of the filter (see Figure \ref{fan_model} for an illustration). The objective of the combined module can be formulated using Equation \ref{eq:fan_loss}.

\begin{align}
\label{eq:fan_loss}
\mathcal{L}(F_{\theta_f}, D_{\theta_d}) &= \lambda \mathbb{E}_{h} || F_{\theta_f} (h) - h ||_2^2 + \\ \nonumber
        & \mathbb{E}_h \bigg[ y\cdot\log(D_{\theta_d}(F_{\theta_f}(h))) + \\ \nonumber
        & (1-y)\cdot\log(1-D_{\theta_d}(F_{\theta_f}(h))) \bigg]
\end{align}

The parameter $\lambda$ is a weight that controls the importance of the first term with respect to the second, and $y$ is the ground truth gender label of the example (a protected attributed). Observe that when we dissect the objective function, we have two distinct terms. The first term represents the reconstruction loss.  The reconstruction loss term is differentiable with respect to the filter parameters $\theta_f$ and it is independent of the discriminator parameters $\theta_d$. The goal is to keep this term approximately at zero in order to attain perfect preservation of the original information. The second term represents cross-entropy. Cross-entropy measures how accurately the discriminator is able to predict the sensitive attribute from the filtered embedding. 

We minimize the combined loss over $\theta_f$ and maximize the combined loss over $\theta_d$ during training.
On the one hand, the filter aims at minimizing the reconstruction loss. On the other hand, the discriminator aims at  minimizing the cross-entropy loss. Intuitively, the optimum saddle point is reached when the discriminator cannot predict the sensitive attribute from the filtered input. The second term of the loss forces the filter to remove the sensitive information from the input embedding, while the first term of the loss (the reconstruction loss) forces the filter to leave the input as much unchanged as possible.

Note that our objective is markedly different from the compositional approach proposed by \cite{bose2019compositional}, where the non-sensitive information is preserved not through the reconstruction loss, but using the edge loss $f_r(h,t)$ coming from the embedding algorithm. The improvement of using the reconstruction is twofold. 
\begin{itemize}
    \item Only the embedding of the entities to filter are required; when using the edge loss, on the other hand, all the triples are necessary to preserve the non-sensitive information.
    \item The reconstruction loss can be used independently of the embedding algorithm used.
\end{itemize}

%% file: 04_experiments_results.tex
\begin{table*}[h]
\centering
\begin{tabular}{ |c|c|c|c|c|c| } 
 \hline
  &  Prediction Task & \textit{TransH} (834 epochs) & \textit{TransH} (999 epochs) & \textit{TransD} (834 epochs) & \textit{TransD} (912 epochs) \\
  \hline
 \multirow{2}{*}{Unfiltered}    & Gender       & 0.67 & 0.67 & 0.68 & 0.68 \\
                                & Occupation   & 0.49 & 0.49 & 0.49 & 0.49 \\
 \hline
 \multirow{2}{*}{$\lambda=0.5$} & Gender       & 0.50 & 0.52 & 0.50 & 0.51 \\
                                & Occupation   & 0.48 & 0.49 & 0.44 & 0.43 \\
                                
 \hline
 \multirow{2}{*}{$\lambda=0.05$} & Gender      & 0.44 & 0.54 & 0.51 & 0.51 \\
                                 & Occupation  & 0.49 & 0.48 & 0.41 & 0.43 \\
 \hline
\end{tabular}
\caption{Results of our debiasing algorithm. The columns represent different embedding algorithms and the number of training epochs. The rows denote three debiasing approaches: unfiltered embeddings and two applications of FAN with different $\lambda$ values. For each embedding algorithm, the top value shows gender prediction accuracy and the bottom value shows the occupation prediction accuracy. Our debiasing algorithm is able to reduce the accuracy of gender prediction to a random event, without hurting the occupation prediction accuracy.}
\label{debias_table}
\end{table*}

\begin{figure*}[h!]
\centering
\begin{subfigure}{.5\textwidth}
    \includegraphics[width=\textwidth]{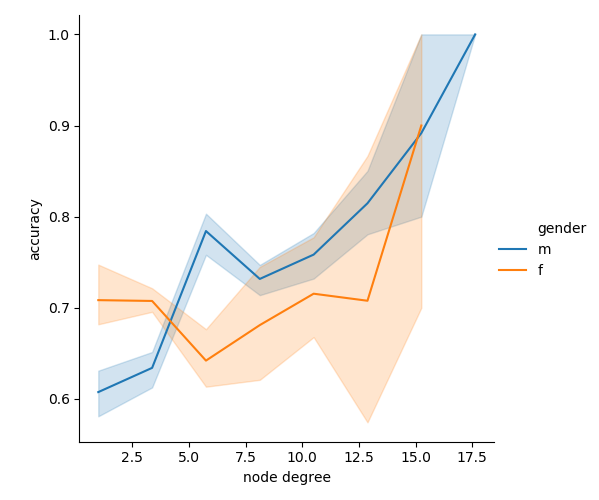}
  \label{fig:gend_acc_unfilt}
\end{subfigure}%
\begin{subfigure}{.5\textwidth}
    \includegraphics[ width=\textwidth]{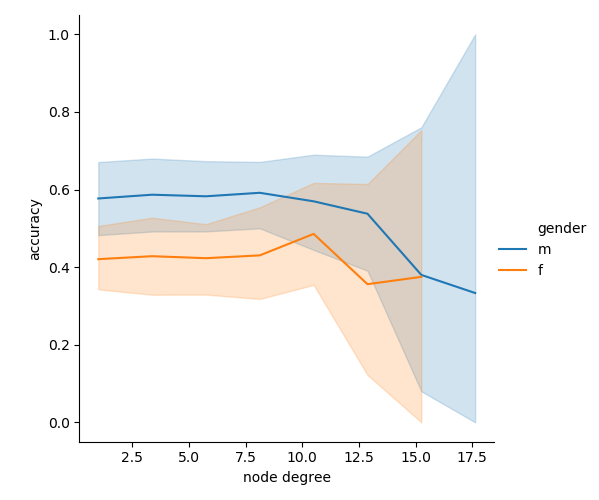}
  \label{fig:gend_acc_filt}
\end{subfigure}%
\caption{Left: gender prediction accuracy against node degree for \textit{unfiltered} embeddings. Right: gender prediction accuracy against node degree for \textit{filtered} embeddings generated by our FAN algorithm. Colored bands represent confidence intervals. Results are obtained starting from embeddings trained with TransH for 999 epochs. We can see that FAN is able to remove gender bias from both high- and low-degree entities.}
\label{gend_acc}
\end{figure*}

In this section we describe the results of the following experiments: (i) exploring popularity bias in network embeddings, (ii) exploring gender bias in KGEs, (iii) debiasing KGEs using our filtering network (FAN).

\subsection{Popularity Bias in Network Embeddings}

\begin{figure}[h!]
\centering
\includegraphics[width=0.5\textwidth]{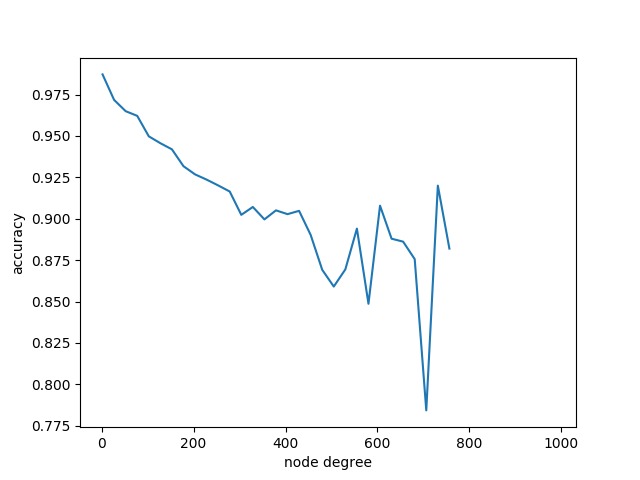}
\caption{Test edge prediction accuracy against node degree for \textit{node2vec} on the \textit{AstroPh} dataset. Degrees are grouped in bins, and mean accuracy is shown within each bin. We see that the prediction accuracy is correlated with node popularity.}
\label{fig:pop_bias_n2v}
\end{figure}

In order to expose the popularity bias, we evaluate the link prediction accuracy of the network embedding using a binary classifier. The classifier simply aims to predict the existence of an edge between two given nodes in the embedding space. The networks used in our experiments are sparse in nature, with the probability of the existence of an edge between two nodes being very low, approximating zero. In order to deal with this skewness and maintain a balance between classes while training, we under sample the negative class. Specifically, for each positive triple $\left(h,r=1,t\right)$, where in general $r \in \left\{0,1\right\}$, we apply negative sampling by replacing the tail entity \textit{t} with a random node. 

Our experiments use a simple Multi-Layer Perceptron (MLP) neural network architecture for binary classification with ReLU activation function.
On the test data, we evaluate the average edge prediction (or link prediction) accuracy for each node. For each node \textit{v}, we consider all the links in which node \textit{v} appears and calculate the prediction accuracy for these edges. 

Figure \ref{fig:pop_bias_n2v} presents the test accuracy against the node degree for \textit{node2vec} on the \textit{AstroPh} dataset, with the degree grouped in bins and the mean accuracy shown within each bin.
Overall, the results indicate that low degree nodes have higher accuracy.
We argue that this is due to the fact that the embedding algorithm performs the same number of random walks for each node, which results in embeddings having more coverage about the topology of the neighborhood of low degree nodes than for high degree nodes.
We see a drop, followed by a rise in edge prediction accuracy around node-degree 700, which warrants further investigation into this phenomenon.

\subsection{DBpedia preprocessing}
DBpedia \cite{auer2007dbpedia} \textemdash \ a crowd-sourced community effort to extract structured content from the information created in various Wikimedia projects \textemdash \ provides a unique research context to examine our questions. 
Structured information curated in DBpedia is available for everyone on the web and resembles an Open Knowledge Graph (OKG). As the DBpedia dataset is extremely sparse, huge, and generally inconsistent, an extensive and rigorous set of preprocessing and subsampling steps were necessary.

After exploratory analysis of the DBpedia graph, we decided to only sample nodes for people in the US, defined as all nodes in the knowledge graph having category ``dbo:Person'' and having any of the following outgoing relations: ``dbp:nationality'', ``dbo:nationality'', ``dbp:country'', ``dbo:country'' with any of the following nodes as tail: ``dbr:United\_States'', ``American'', ``United States''. 

First, we consider all incoming and outgoing relations for all US people, leading to a sub-graph with about 10 millions triples. This sample was then used to train the embedding. However, this method was not able to capture the relations properly as the dataset was mainly composed by few non-semantic relations (about 5 millions triples came from relations like ``dbo:wikiPageWikiLink'', ``rdf:type'', ``dct:subject''), which resembled the characteristics of a normal network as compared to a KG, and completely warped the geometry of the embeddings.

In a second attempt, we identified and removed the most-frequent non-semantic relations and also all relations that appears in only 10 triples or less, as we noticed that they were very noisy. The resulting sample consisted of about 2 millions triples. Although this dataset performed better than the previous sample in embedding learning, the performance was not par with common baselines of these algorithms.
After carefully examining the results, we identified that this lack of performance was mainly due to the crowd-sourced nature of the data. The nomenclature of the nodes, and even of the relations, were incredibly inconsistent. For example, relations and nodes with the same meaning were given many different notations. To explain this issue, take the occupation relation as an example. To express the concept of occupation, there exist multiple relations: ``dbo:occupation'', ``dbo:profession'', ``dbr:occupation'', and ``dbr:profession''. Some of these relations point to nodes which take the form of occupations \textit{URI} (e.g. ``dbo:Writer''), while others take the form of strings (e.g. ``writer''), and finally a huge chunk of them points to dummy nodes, which replicates the person name, which in turn have a \textit{title} relation pointing to the actual occupations.

Additionally, we often had a string containing a list of occupations as tail without a consistent separator, and hence we had the problem of synonyms.
Therefore, it became clear that using simple raw, unprocessed values would be ineffective in capturing semantically meaningful concepts, because all these variations of the same occupation will be considered different entities.

\begin{figure*}[h!]
\includegraphics[width=\textwidth]{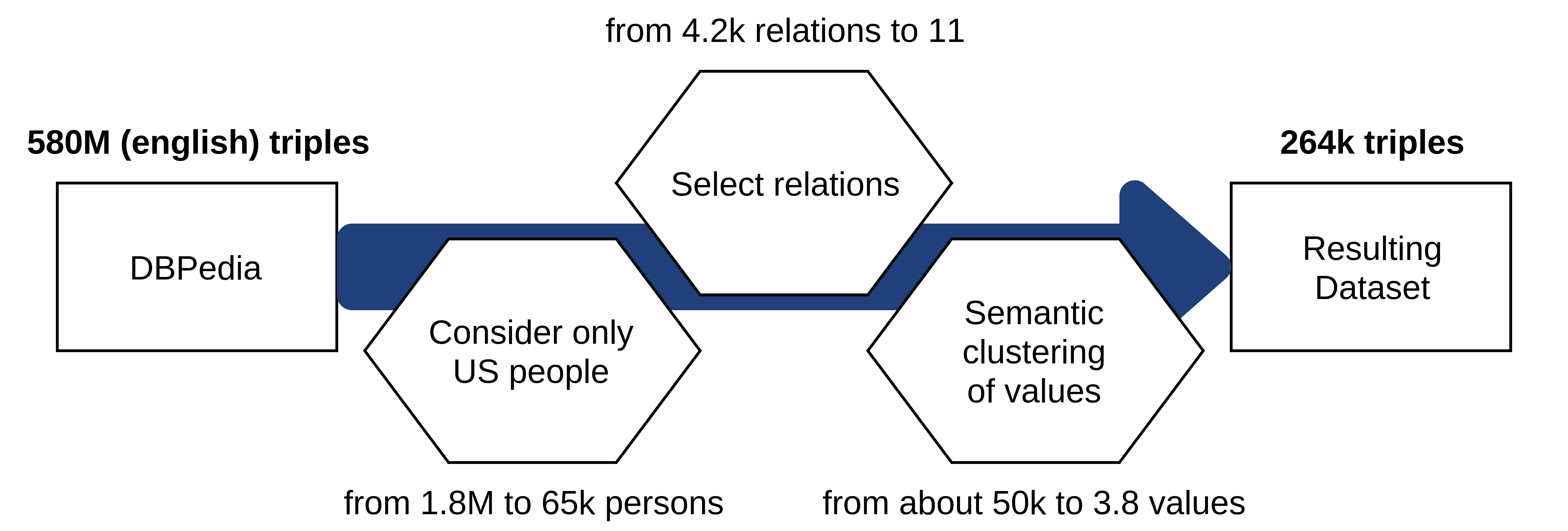}
\caption{DBpedia dataset preprocessing.}
\label{fig:dbpedia_processing}
\end{figure*}

Finally, as a third attempt we select $11$ meaningful relations and individually looked, parsed and cleaned each of them, manually grouping nodes referring to the same concept and cleaning the errors we found in the dataset (e.g. conferences or a submarine classified as a ``dbo:Person''). This last version of \textit{DBpedia} has about 200k triples and 44 occupations, leading to fast training and meaningful embeddings.
We summarize our dataset preparation steps in Figure~\ref{fig:dbpedia_processing}.

\subsection{Debiasing KGE for Gender Bias using FAN}

We considered four different embeddings for this experiment: TransH trained for 834 and 999 epochs and TransD trained for 834 and 912 epochs.
To train the FAN, for each of them, we pretrain a filter, that aims to learn an identity mapping of the embedding, and a discriminator, aiming to predict the gender, separately, for 10 epochs. 

The adversarial training is initiated by jointly training the filter and the discriminator, running one training step for the filter every five steps for the discriminator. Both the filter and the discriminator are implemented as MLP with one hidden layer for the filter and two for the discriminator, Leaky ReLU activation function and dropout rate of 0.5 for non output layers. We then use the learned filter to train two discriminators, to predict gender and occupation from the filtered embeddings.

We present our cross-validated results in Table \ref{debias_table}. Observe that we are able to remove the gender information from the embeddings, making it impossible for the classifier to predict the gender, while at the same time keeping constant performance for occupation prediction.

Furthermore, we evaluated the prediction accuracy in gender prediction against the node degree. Figure \ref{gend_acc} displays this for both the unfiltered and the filtered embeddings.
For the unfiltered embeddings, we observe that the classifier can predict the gender and the performance improves for high-degree nodes, indicating that the gender information is not only contained in the explicit gender relation. The observation of popularity bias is opposite of what we observe in network embeddings (Figure \ref{fig:pop_bias_n2v}), showing the additional challenge in the case of KGs. For the filtered embeddings, the classifier is not able to correctly predict the gender, achieving near-random prediction accuracy. 

These results show that our filtering network FAN is able to remove gender biases from the KGEs, from both high degree and low-degree entities.

%% file: 05_conclusion.tex
In this presentation of our \textbf{work-in-progress}, we describe popularity bias in network embeddings and we explore the presence of gender bias in KGEs. We also describe FAN, a new algorithm for debiasing KGEs. Our expeirmental results suggest that FAN is able to remove gender bias in KGs, for both high- and low-degree nodes. In other words, it can deal with both popularity and gender bias in KGs.

The FAN framework could also be useful in other applications. Future works should explore the FAN framework in further detail, by applying it to different tasks than knowledge graph embedding debiasing. In fact, the objective presented in Equation \ref{eq:fan_loss} is independent of the specific task, and therefore in principle FAN can be applied whenever the task requires learning to filter out specific information, while retaining as much as possible of the remaining information. Our work also has some limitations, which we would like to address in future work. First, we would like to further explore popularity biases in network embeddings, with more datasets and embedding algorithms. Second, for KGEs, we would also like to explore other types of biases, and experiment on more datasets and embedding algorithms.